\documentclass{article}
\usepackage{amssymb}
\usepackage{wrapfig}
\usepackage[preprint]{corl_2026} 
\usepackage{amsmath}
\usepackage{xcolor}
\usepackage{amsfonts}
\usepackage{graphicx}
\usepackage{float}
\usepackage{enumitem}
\usepackage{booktabs}
\usepackage{multirow}
\usepackage{pgfplots, tikz}
\usepackage{graphicx}
\usepackage{twemojis}
\pgfplotsset{compat=1.18}

\newcommand{\approach}{$\pi\mathbf{R}^2$}

\title{\approach: Reactive Real-time Flow Policies}

\author{
  Sungjae Park, Shubham Tulsiani\\
  Carnegie Mellon University \\
  \texttt{\{sungjae2, stulsian\}@andrew.cmu.edu} \\
}
\usepackage{lib}
\usepackage{booktabs}
\usepackage{wrapfig}
\usepackage{subcaption}
\usepackage{subcaption}
\usepackage{graphicx}
\usepackage{algorithm}
\usepackage{algpseudocode}

\begin{document}
\maketitle

\begin{abstract}
Generalist manipulation policies increasingly take the form of action-chunking flow policies built on large pretrained backbones. Such chunks run open-loop, so the policy cannot react to sensory input arriving mid-execution, sacrificing \emph{reactivity}. Replanning more often would restore it, but the perception-to-action pipeline (a large backbone plus multiple denoising steps) is too slow: this \emph{latency} forbids frequent replanning and leaves committed actions stale, making such policies ill-suited for dynamic, closed-loop control. We present~\approach{}, which makes these policies reactive and real-time while retaining large backbones, expressive multi-modal policies, and multi-action prediction. Built on the per-position noise schedule of diffusion forcing, \approach{} contributes two ideas. First, it splits conditioning into a fast channel (proprioception, fresh every tick) and an asynchronously updated slow channel (vision-language features), so the policy reacts to proprioception within a chunk while tolerating stale vision. Second, a latency-adaptive flow schedule treats in-flight actions as inpainting conditioning and emits actions in one denoising step per call, letting one trained model adapt to varying hardware latency. Requiring minimal modification to existing architectures, \approach{} can be finetuned from a pretrained policy: applied to GR00T-N1.7 on a real xArm6+XHand platform, it replans closed-loop roughly $4\times$ faster than the base policy ($25$~Hz on an A5000 GPU), acting on a fresh observation every $40$~ms. Across simulation and real-world manipulation tasks, \approach\ improves the success rate by up to $23\%$ in simulation and $30\%$ in the real world over the strongest baseline.

\textbf{Project page:} \href{https://pi-r2-flow.github.io/}{https://pi-r2-flow.github.io/}\end{abstract}

\keywords{Reactivity, Real-time Inference, Flow Policies, Robot Foundation Models}

\section{Introduction}
\label{sec:introduction}

We have witnessed remarkable progress in training `robotics foundation models'~\cite{pi0, pi05, bjorck2025gr00t, rt1, rt2, openvla}, and three design choices have become ubiquitous across recent efforts. First, \textbf{large pre-trained backbones} such as vision-language models (VLMs)~\cite{liu2023visual, alayrac2022flamingo, bai2025qwen3, team2023gemini, beyer2024paligemmaversatile3bvlm, deitke2024molmopixmoopenweights} are leveraged to process visual (and language) input and extract rich, generalizable representations for downstream action prediction. Second, \textbf{expressive policy architectures} such as diffusion~\cite{chi2023diffusion} and flow matching~\cite{generalflow, pan2025much} allow effectively capturing the multi-modal action distributions inherent in diverse human demonstrations. Finally, \textbf{action chunking}~\cite{zhao2023learning, zhang2025action} enables these models to jointly predict multiple actions, providing a richer training signal and improving the temporal consistency of executed actions.

\begin{figure}[t]
    \centering
    \includegraphics[width=0.75\linewidth]{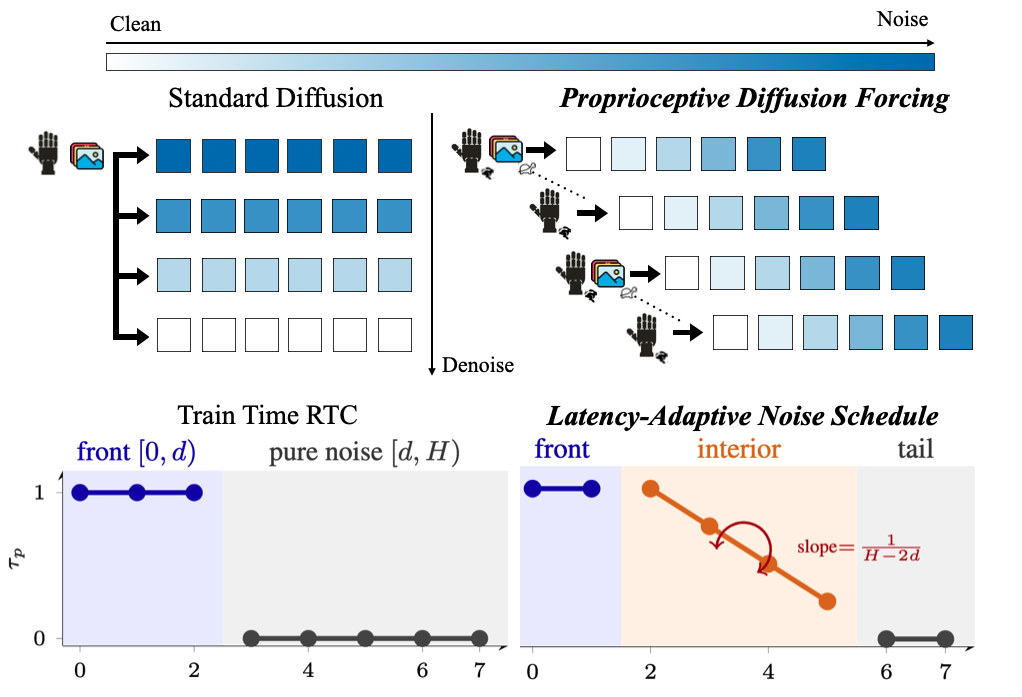}
    \caption{Overview of~\approach. \textbf{Top.} While standard diffusion/flow matching relies on stale observations to predict actions via iterative denoising,~\approach~disentangles the observation into a fast channel (proprioception) and a slow channel (image encoding, VLM embedding, etc.), and uses up-to-date observations for each denoising step. \textbf{Bottom.} To incorporate latency and smooth execution, we adopt an adaptive noise schedule, where the action chunk is divided into three regions: clean actions to be taken during inference, actions with increasing noise level following Diffusion Forcing~\cite{chen2024diffusion}, and pure noise with the same length as clean actions. Compared to Train-Time RTC, this enables faster inference and smoother actions.}
    \label{fig:concept}
    \vspace{-16pt}
\end{figure}

On the one hand, these choices have been crucial in unlocking generalizable and scalable imitation learning. However, they result in \emph{limited reactivity} and \emph{increased latency}, making these foundation models ill-suited for dynamic manipulation tasks that demand fast, smooth, and reactive control. Specifically, the execution of a `chunk' of predicted actions is performed open-loop, without the ability to react to the sensory input streaming in. While the reactivity can in theory be improved by only executing a small `sub-chunk' (in the limit just one action) and re-predicting with updated sensory input, naively doing so is infeasible due to the significant computation required in the `perception-to-action' prediction -- a large visual (and language) processing backbone followed by multiple denoising iterations for flow matching policy inference. This computational bottleneck coupled with action chunking flow matching thus results in two (related) fundamental limitations: a) reduced reactivity, as a large (sub-)chunk of actions must be executed open-loop while the next chunk is predicted, and b) increased latency, as the predicted actions are a function of `stale' sensory input.

In this work, we develop \approach: a framework that allows reactive real-time policy inference \emph{without} compromising on the design principles of large pre-trained backbones, expressive multi-modal policies, and multi-action prediction. We note that instead of flow (or diffusion) policy inference, which maps uniformly random noise to action chunks over multiple denoising iterations, leveraging the flexible schedule in `diffusion forcing'~\cite{chen2024diffusion} offers several benefits: a) the immediate actions to be executed can be at a lower noise level and predicted in fewer iterations, and b) different denoising iterations involved in predicting an action can rely on updated sensory input as available. Although prior work has explored diffusion forcing for policy learning~\cite{hoeg2024streaming}, we address the key issues that prevent its widespread adoption for robotics foundation models.

First, while the (immediate) action denoising requires less computation, the backbone (\eg VLM) feature processing becomes the bottleneck for `perception-to-action prediction' latency. Our key insight that allows us to circumvent this bottleneck is that not all input modalities need to be processed at the same rate: proprioceptive signals (joint positions, velocities, torques, and contact forces) can be retrieved and processed orders of magnitude faster than images or text. Moreover, for dynamic tasks, proprioception carries sufficient information for \emph{local} reactive corrections (\emph{e.g.}, responding to an unexpected contact or compensating for object slip), while vision and language provide the \emph{global} context needed to guide coarse motion plans. We thus disentangle the conditioning for action denoising into asynchronously updated `slow' and always up-to-date `fast' features,  effectively allowing high-frequency proprioceptive control within an action chunk.
Second, despite our asynchronous diffusion forcing, practical constraints (communication delays, GPU bandwidth) often prevent zero latency. We develop a delay-adaptive noise schedule for diffusion forcing that enables seamless real-time execution despite latency, allowing the model to continue executing the `previous' actions while ensuring smooth temporally coherent outputs from subsequent denoising iterations. 

Together, these design choices make \approach{} roughly $4\times$ faster than the base policy for closed-loop \emph{replanning}, achieving 25~Hz with a large-scale VLA (GR00T N1.7~\cite{bjorck2025gr00t}) on A5000 GPUs. Crucially, this means the policy predicts each action using fresh observation every 40~ms, rather than committing to a long action chunk and replanning only sparsely (\eg at $\sim$7~Hz). \approach{} thus matches the reactivity of simple feedforward policies while retaining the expressivity of flow matching, the training benefits of chunk-based inference, and the semantic grounding of pretrained backbones. We first validate \approach{} in simulation where we highlight the benefits of increased reactivity and reduced latency compared to the typical action chunking flow policies. We also show how, instead of training from scratch, we can finetune pre-trained VLAs with a diffusion forcing schedule and demonstrate the efficacy of \approach{} for complex real-world manipulation tasks, where we find that it improves over the strongest baseline by up to $23\%$ in simulation and $30\%$ in the real world.

\vspace{-12pt}

\section{Related Work}
\label{sec:related_work}

\vspace{-2mm}
\paragraph{Action Chunking and Flow Policies.} Predicting a \emph{chunk} of future actions rather than a single action, introduced by ACT~\citep{zhao2023learning}, improves the temporal coherence of imitation-learned policies and has since become standard, including for mobile and contact-rich manipulation~\citep{fu2024mobilealohalearningbimanual, zhao2024aloha}. This design has been widely adopted by expressive multi-modal policies based on diffusion~\citep{chi2023diffusion} and flow matching~\citep{zhang2024flowpolicyenablingfastrobust, yan2025maniflowgeneralrobotmanipulation}. For dynamic tasks, however, committing a (sub-)chunk open-loop is suboptimal, as it cannot react to sensory input arriving mid-execution. While our work also reasons over a chunk of actions, its use of diffusion forcing predicts the immediate actions with lower latency and restores reactivity by conditioning successive denoising iterations on progressively fresher sensory input.

\vspace{-2mm}
\paragraph{Generalist Manipulation Policies.} Foundation robotics models also couple chunked flow-matching action heads with large vision-language backbones, yielding vision-language-action (VLA) models that scale imitation learning across diverse tasks and embodiments~\citep{open_x_embodiment_rt_x_2023, khazatsky2024droid, rt2, kim2024openvla}. Notable examples include $\pi_0$/$\pi_{0.5}$ and the dual-system GR00T~N1, whose vision-language module feeds a flow-matching action head conditioned on robot proprioception~\citep{pi0, pi05, nvidia2025gr00tn1openfoundation}. While this has yielded impressive, generalizable policies, coupling a large backbone with multi-step flow-matching denoising only exacerbates the reactivity problem: the combined per-call latency limits how often the policy can replan. Crucially, this backbone latency would persist even under a diffusion-forcing schedule, since each emitted action still requires a full forward pass through the slow semantic backbone. We instead exploit an asymmetry these models already expose---proprioception is far cheaper to process than vision and language---refreshing proprioceptive conditioning at every control tick while letting the slow vision-language features update asynchronously, making this same class of models reactive.

\vspace{-12pt}
\paragraph{Real-time Execution of Action-Chunking Policies.}
The high inference latency of modern VLAs has motivated methods for executing action chunks smoothly while the next prediction is computed~\cite{xie2026dynamicvla}. RTC maintains continuity across asynchronous chunks via inference-time inpainting on a frozen action prefix, while Training-Time RTC instead folds this prefix conditioning into training, avoiding the extra inference-time cost~\citep{black2026real, black2025training}. Streaming Diffusion Policy and Streaming Flow Policy emit actions incrementally from a diffusion-forcing-style variable-noise buffer maintained across observations~\citep{hoeg2024streaming, jiang2025streamingflowpolicysimplifying}. None, however, address the latency of repeatedly conditioning on a large semantic backbone; Streaming Diffusion Policy, for instance, uses compact visuomotor policies and runs synchronously.
Concurrent to our work, FASTER~\citep{lu2026fasterrethinkingrealtimeflow} similarly reduces reaction latency in flow-based VLAs via a diffusion-forcing-inspired horizon-aware schedule with action-prefix conditioning, but forwards the backbone once per chunk and conditions all actions on a single fixed observation, improving latency and smoothness without improving reactivity to sensory input arriving during execution. \approach{} instead refreshes proprioceptive conditioning within the chunk, reacting to incoming feedback as actions execute while vision-language features update asynchronously.

\vspace{-12pt}

\section{\approach}
  \label{sec:method}
  \vspace{-10pt}

  Our goal is to modify existing large-scale flow policies for higher reactivity and lower inference latency. We first review the building blocks -- flow matching and diffusion forcing -- and
  then introduce two modifications that together yield a real-time, closed-loop flow policy. The modifications require only a flow matching action head conditioned on a heavy pretrained backbone, so they apply equally to world-action models~\cite{ye2026world, kim2026cosmos, li2025unified, zhu2025unified}; we instantiate and evaluate on VLAs, the most common member of this family.

  \subsection{Preliminary}
  \label{sec:prelim}

  \subsubsection{Flow Matching and Action Chunking Policy}
  \label{sec:prelim_flow}

  We build on flow matching~\cite{lipman2022flow, zhang2024affordance} with action chunking~\cite{zhao2023learning, zhang2025action}. Flow matching learns a velocity field
  $v_\theta(\mathbf{x}_t, t)$ that transports samples from noise $p_0 = \mathcal{N}(\mathbf{0}, \mathbf{I})$ to data $p_1 = p_\mathrm{data}$ along the conditional interpolation
  $\mathbf{x}_t = (1{-}t)\,\boldsymbol{\epsilon} + t\,\mathbf{x}_1$, $t \in [0, 1]$, trained against the conditional velocity $u_t(\mathbf{x}_t \mid \mathbf{x}_1) = \mathbf{x}_1 -
  \boldsymbol{\epsilon}$:
  \begin{equation}
  \mathcal{L}_\mathrm{FM} = \mathbb{E}_{t, \mathbf{x}_1, \boldsymbol{\epsilon}}\!\left[\|v_\theta(\mathbf{x}_t, t) - (\mathbf{x}_1 - \boldsymbol{\epsilon})\|^2\right].
  \end{equation}
  In the action-chunking policy setting, $\mathbf{x}_1 = (\mathbf{a}_t, \dots, \mathbf{a}_{t+H})$ is a chunk of $H$ future actions conditioned on observation $\mathbf{o}_t$, generated
  jointly by $K$ denoising steps; the robot executes the first $h \le H$ actions before re-planning. \textbf{Crucially, all $H$ positions share a single noise level $t$ and the chunk is
  committed open-loop: the $h$ executed actions never account for new observations during their execution window}, limiting the reactivity that manipulation tasks demand.

  \subsubsection{Diffusion Forcing and Streaming Diffusion}
  \label{sec:prelim_df}

  Diffusion forcing~\cite{chen2024diffusion, song2025history} generalizes flow matching by assigning each chunk position $p \in \{0, \dots, H{-}1\}$ an independent noise level $\tau_p \in
  [0, 1]$:
  \begin{equation}
  \mathbf{x}_{\tau, p} = (1{-}\tau_p)\,\boldsymbol{\epsilon}_p + \tau_p\,\mathbf{a}_p,
  \end{equation}
  and the model $v_\theta(\mathbf{x}_\tau, \boldsymbol{\tau}, \mathbf{o})$ predicts a per-position velocity. The flexibility of position-dependent noise levels enables \emph{closed-loop}
  action prediction by conditioning successive denoising steps on progressively more recent observations within a single chunk. \emph{Streaming diffusion}~\cite{hoeg2024streaming} is one
  such instantiation: it imposes a linearly increasing noise schedule across the chunk so that the leading positions are fully denoised within a few steps, and each denoising step
  incorporates a fresh observation.

  In this work, we also adopt diffusion forcing, focusing on an underexplored regime: enabling large-scale flow policies (\eg VLAs) to be more reactive and closed-loop, where large latency naturally arises
  from its model size and components. The following two subsections present the architectural and scheduling modifications that make diffusion forcing a practical real-time, reactive policy on top
  of any flow matching action head.

  \subsection{Proprioception-Reactive Diffusion Forcing}
  \label{sec:async-vlm}
  A VLA processes the observation $\mathbf{o}_t = (\mathbf{s}_t, \mathbf{I}_t, \mathbf{T}_t)$ -- proprioception $\mathbf{s}_t$ (joint positions, end-effector pose, \etc), image $\mathbf{I}_t$,
  and language $\mathbf{T}_t$ -- to predict an action chunk $(\mathbf{a}_t, \dots, \mathbf{a}_{t+H})$. Internally, the inputs are preprocessed (dominated by image processing) and passed
  through the large backbone(\eg VLM), producing language-aligned features; proprioception is processed by a small MLP into a state embedding; the DiT action head then conditions on the concatenated
  representation and runs $K$ denoising steps over the chunk. In practice, these components differ greatly in cost. On RTX A6000 with GR00T-N1.7, image preprocessing + VLM ($\sim 60$~ms) plus DiT denoising ($K{=}4$ steps, $\sim 80$~ms) sum to
  $\sim 140$~ms per call -- $\sim 7$ control ticks at $50$~Hz. This motivates disentangling the slow VLM path from the fast action prediction loop.

  We propose to disentangle the DiT's conditioning (Fig.~\ref{fig:concept}, left) into a \emph{slow channel} (vision and language features from the VLM and image/text encoders) and a
  \emph{fast channel} (proprioception, fresh every tick). This split mirrors the structure of manipulation itself: vision and language provide coarse spatial and task guidance, while
  fresh proprioception drives the fine motor refinement that precision demands. Diffusion forcing additionally reduces the per-call denoising work: because the front of the chunk is kept
  near $\tau = 1$, a single denoising step per call suffices to produce clean actions at the front (one or more, depending on the schedule; see Sec.~\ref{sec:slope}), versus $K$ steps in
  standard chunked denoising. Combining the two, the DiT action head runs one denoising step against fresh proprioception and a \emph{cached} slow feature; vision and language are
  processed asynchronously in a background thread (image preprocessing + VLM forward) and refresh the cache when complete, so the action head conditions on a slow feature of age
  $d_\mathrm{vlm}$ ticks (the vision/text wall-clock latency). Per-call cost of the action head is therefore one NFE only, independent of backbone size.

  To absorb the resulting slow-channel staleness, we delay the slow channel at training time by $d_\mathrm{vlm} \sim \mathrm{Uniform}\{0, \dots, d_\mathrm{vlm}^{\max}\}$ and supply the
  delay value through a learned embedding indexed by the integer delay, added to the slow representation; at deploy time the measured $d_\mathrm{vlm}$ uses the same embedding. The resulting policy is
  \emph{proprioception-reactive}: it reacts to fresh joint state every denoising step while tolerating bounded staleness on vision and language, recovering closed-loop control even behind
  a large VLA backbone. We refer to this architecture variant as \emph{Proprioception-Reactive Diffusion Forcing}.

  \subsection{Latency-Adaptive Flow Schedule}
  \label{sec:slope}

  By combining diffusion forcing with asynchronous vision/text processing and single-step DiT inference, the per-call delay $d$ of the action prediction loop (distinct from the vision-language staleness $d_\mathrm{vlm}$ of Sec.~\ref{sec:async-vlm}) reduces to one denoising step
  of the action head. However, in reality, $d$ varies: it depends on each component size and inference GPU, includes network latency when policy and robot run on separate machines (e.g., Ethernet round-trips in remote-inference setups), and fluctuates with computational load and per-call
  jitter. The policy must therefore handle a range of inference delays $d \in [1, d_\mathrm{max}]$.

  A naive linearly increasing noise schedule, as in standard streaming diffusion~\cite{hoeg2024streaming}, does not address this regime: it implicitly assumes $d = 0$ (instantaneous inference, no in-flight
  actions), so when applied at $d > 0$ the returning chunk can jump at the chunk boundary~\cite{black2026real, tang2025vlash}. We address both issues with a per-position noise schedule
  parameterized by the inference delay $d$: a clamped-clean front carries the $d$ in-flight actions as inpaint conditioning, a ramped interior emits $d$ clean actions per call at the cost
  of one denoising step, and randomizing $d$ at training lets a single model adapt to whatever value is measured at each call.

  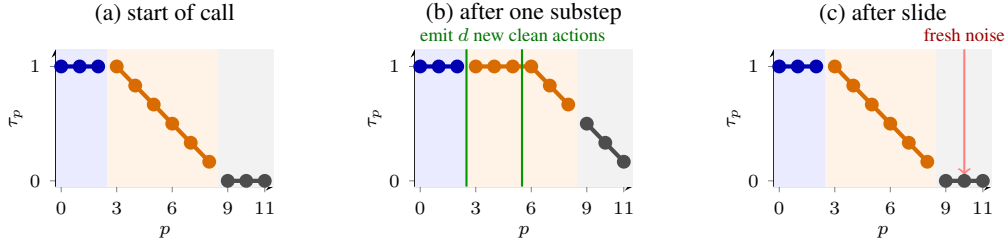
\begin{figure}[t]   
  \centering
  \pgfplotsset{
    schedplot/.style={
      width=0.32\linewidth, height=3.4cm,
      xmin=-0.3, xmax=11.5, ymin=-0.05, ymax=1.15,
      xlabel={$p$}, xlabel style={font=\scriptsize, yshift=2pt},
      ylabel={$\tau_p$}, ylabel style={font=\scriptsize, yshift=-4pt},
      xtick={0,3,6,9,11}, ytick={0,1},
      ticklabel style={font=\scriptsize},
      axis lines=left, clip=false,
    }
  }
  \begin{tikzpicture}
  \begin{scope}
  \begin{axis}[schedplot, title={\small (a) start of call}]
    \fill[blue!8]    (axis cs:-0.3,-0.05) rectangle (axis cs:2.5,1.15);
    \fill[orange!10] (axis cs:2.5,-0.05)  rectangle (axis cs:8.5,1.15);
    \fill[gray!10]   (axis cs:8.5,-0.05)  rectangle (axis cs:11.5,1.15);
    \addplot[ultra thick, blue!70!black, mark=*, mark size=1.8pt]
      coordinates {(0,1) (1,1) (2,1)};
    \addplot[ultra thick, orange!85!black, mark=*, mark size=1.8pt]
      coordinates {(3,1) (4,0.833) (5,0.667) (6,0.5) (7,0.333) (8,0.167)};
    \addplot[ultra thick, gray!60!black, mark=*, mark size=1.8pt]
      coordinates {(9,0) (10,0) (11,0)};
  \end{axis}
  \end{scope}
  \begin{scope}[xshift=0.34\linewidth]
  \begin{axis}[schedplot, title={\small (b) after one substep}]
    \fill[blue!8]    (axis cs:-0.3,-0.05) rectangle (axis cs:2.5,1.15);
    \fill[orange!10] (axis cs:2.5,-0.05)  rectangle (axis cs:8.5,1.15);
    \fill[gray!10]   (axis cs:8.5,-0.05)  rectangle (axis cs:11.5,1.15);
    \addplot[ultra thick, blue!70!black, mark=*, mark size=1.8pt]
      coordinates {(0,1) (1,1) (2,1)};
    \addplot[ultra thick, orange!85!black, mark=*, mark size=1.8pt]
      coordinates {(3,1) (4,1) (5,1)(6,1)(7,0.833) (8,0.667) };
    \addplot[ultra thick, gray!60!black, mark=*, mark size=1.8pt]
      coordinates {(9,0.5) (10,0.333) (11,0.167)};
    \node[font=\scriptsize, green!45!black] at (axis cs:5.0, 1.28) {emit $d$ new clean actions};
    \draw[-, green!60!black, thick] (axis cs:5.5,1.15) -- (axis cs:5.5,-0.05);
    \draw[-, green!60!black, thick] (axis cs:2.5,1.15) -- (axis cs:2.5,-0.05);

  \end{axis}
  \end{scope}
  \begin{scope}[xshift=0.68\linewidth]
  \begin{axis}[schedplot, title={\small (c) after slide}]
    \fill[blue!8]    (axis cs:-0.3,-0.05) rectangle (axis cs:2.5,1.15);
    \fill[orange!10] (axis cs:2.5,-0.05)  rectangle (axis cs:8.5,1.15);
    \fill[gray!10]   (axis cs:8.5,-0.05)  rectangle (axis cs:11.5,1.15);
    \addplot[ultra thick, blue!70!black, mark=*, mark size=1.8pt]
      coordinates {(0,1) (1,1) (2,1)};
    \addplot[ultra thick, orange!85!black, mark=*, mark size=1.8pt]
      coordinates {(3,1) (4,0.833) (5,0.667) (6,0.5) (7,0.333) (8,0.167)};
    \addplot[ultra thick, gray!60!black, mark=*, mark size=1.8pt]
      coordinates {(9,0) (10,0) (11,0)};
    \node[font=\scriptsize, red!60!black] at (axis cs:10, 1.28) {fresh noise};
    \draw[->, red!50, thick] (axis cs:10,1.15) -- (axis cs:10,0.05);
  \end{axis}
  \end{scope}
  \end{tikzpicture}
  \caption{\textbf{Inference cycle, one call.} (a) Buffer at $\tau^{\star,d}$ at the start of the call. (b) One Euler substep (one NFE) shifts the schedule right by $d$
  slots: positions $[d, 2d)$ reach $\tau{=}1$ and are released, and the ramp extends through the back (c) The buffer slides $d$ positions: just-emitted
  actions become the new front conditioning, $d$ fresh-noise slots ($\tau{=}0$) are appended -- the schedule reproduces exactly.}
  \label{fig:cycle}
  \vspace{-12pt}
  \end{figure}

  \paragraph{Staircase schedule (Fig.~\ref{fig:concept}, right).}
  For a target inference delay $d$, \approach\ uses the three-region staircase $\boldsymbol{\tau}^{\star,d} \in [0,1]^H$
  \begin{equation}
  \tau^{\star,d}_p =
  \begin{cases}
  1                              & 0 \le p < d \\[2pt]
  1 - \dfrac{p - d}{H - 2d}      & d \le p < H - d \\[6pt]
  0                              & H - d \le p \le H - 1
  \end{cases}
  \end{equation}
  with interior slope $s = 1/(H{-}2d)$. Three roles by position: \textbf{front} $[0, d)$, the in-flight actions, clamped clean as inpaint conditioning; \textbf{interior} $[d, H{-}d)$, a
  linear ramp from clean to noise; \textbf{tail} $[H{-}d, H)$, the $d$ pure-noise slots appended at the back per cycle. This staircase can be viewed as a diffusion-forcing generalization
  of training-time RTC~\cite{black2025training}: both clamp a clean front of $d$ in-flight actions for inpaint conditioning, but RTC applies a single shared noise level over the remaining
  $H - d$ positions, whereas the staircase replaces it with a ramped interior plus a pure-noise tail to enable single-step emission under diffusion forcing.

  \paragraph{Training.}
  At each batch we sample $d \sim \mathrm{Uniform}\{1, \dots, d_\mathrm{max}\}$ and build $\boldsymbol{\tau}^{\star,d}$. The front $d$ slots are filled by the ground-truth actions and
  excluded from the loss via $m_p = \mathbf{1}[p \ge d]$ (mirroring inference-time inpaint conditioning); the interior and tail are noised at the staircase levels and contribute to the
  per-position MSE. We additionally apply symmetric jitter $\tau_p \leftarrow \mathrm{clip}(\tau_p + \delta_p, 0, 1)$ with $\delta_p \sim \mathrm{Uniform}[-j, j]$ to absorb small
  deviations from the central staircase that arise from per-call $d$ variation. With probability $p=0.2$, we instead train on a standard flow schedule (single $\tau \sim
  \mathrm{Uniform}[0,1]$ shared across positions, no mask), so the same network can also denoise a full chunk from pure noise, used at inference to warm-start the buffer at episode start.

  \textbf{Compatibility with existing flow policy action heads.} The procedure above plugs into any DiT-style flow matching action with a one-line architectural change: the AdaLN conditioning (which
  modulates each DiT block by the noise level $\tau$) becomes per-position, with one $(\gamma_p, \beta_p)$ pair per chunk position rather than one shared across positions. All other
  components (attention, MLP, pretrained large backbone, slow/fast feature paths) remain unchanged, so existing pretrained policies, such as VLAs, can be fine-tuned with \approach\ without touching the backbone.

  \paragraph{Inference cycle (Fig.~\ref{fig:cycle}).}
  At episode start we warm up the buffer with standard-flow inference (denoising all $H$ positions from pure noise) and re-noise to match $\boldsymbol{\tau}^{\star,d}$ for the initial
  measured $d$. From then on, each call applies one denoising step that updates each position by
  \begin{equation}
  \mathbf{x}_p \;\leftarrow\; \mathbf{x}_p + \Delta\tau_p \cdot v_\theta(\mathbf{x}, \boldsymbol{\tau}, \mathbf{o})_p,
  \qquad
  \tau_p \;\leftarrow\; \tau_p + \Delta\tau_p,
  \label{eq:euler-step}
  \end{equation}
  with region-dependent per-position advances $\Delta\tau_p$ chosen so the schedule shifts right by $d$ slots: positions $[d, 2d)$ reach $\tau{=}1$ and are emitted, and the rest of the
  buffer rotates forward (Fig.~\ref{fig:cycle}b). The buffer then slides by $d$ positions, with $d$ fresh-noise slots ($\tau{=}0$) appended at the back (Fig.~\ref{fig:cycle}c). The
  schedule reproduces exactly when $d$ holds steady; when $d$ changes between calls the $\Delta\tau_p$ pattern adapts to the new measured $d$, and the buffer is pulled toward the new
  $\boldsymbol{\tau}^{\star,d}$ over a few calls.

\vspace{-5pt}
\section{Experiments}
  \label{sec:experiments}

  \subsection{Simulation Experiments}
  \label{sec:sim}

\textbf{Setup.}
  We first test~\approach~on the Leap Cube Reorientation task in MuJoCo Playground~\cite{zakka2025mujoco}, a common task that requires reactivity to keep the cube from falling and to
  achieve the target cube pose. Demonstrations are collected at $50$~Hz from $4$ RL experts (different seeds), yielding $200$ trajectories total. All methods are trained as state-based
  flow policies, with prediction chunk length $H=16$. We run two studies: an execution-horizon $h$ sweep under zero inference delay (Sec.~\ref{sec:sim_zero}), and a deployment study under a fixed latency budget
  (Sec.~\ref{sec:sim_delay}).

  \begin{figure}[t]
    \centering
    \begin{subfigure}[t]{0.48\columnwidth}
      \centering
      \includegraphics[width=\linewidth]{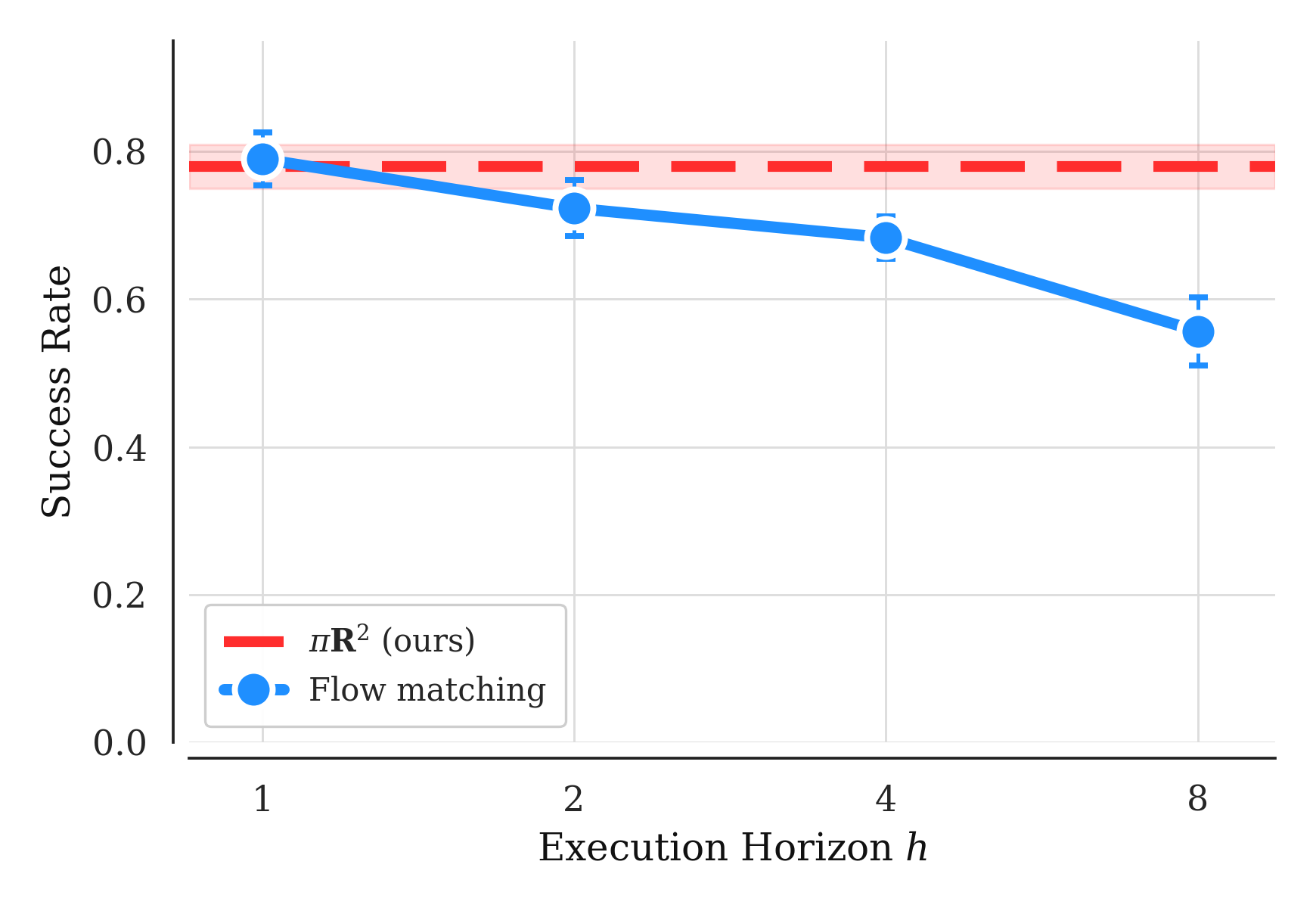}
    \end{subfigure}
    \hfill
    \begin{subfigure}[t]{0.48\columnwidth}
      \centering
      \includegraphics[width=\linewidth]{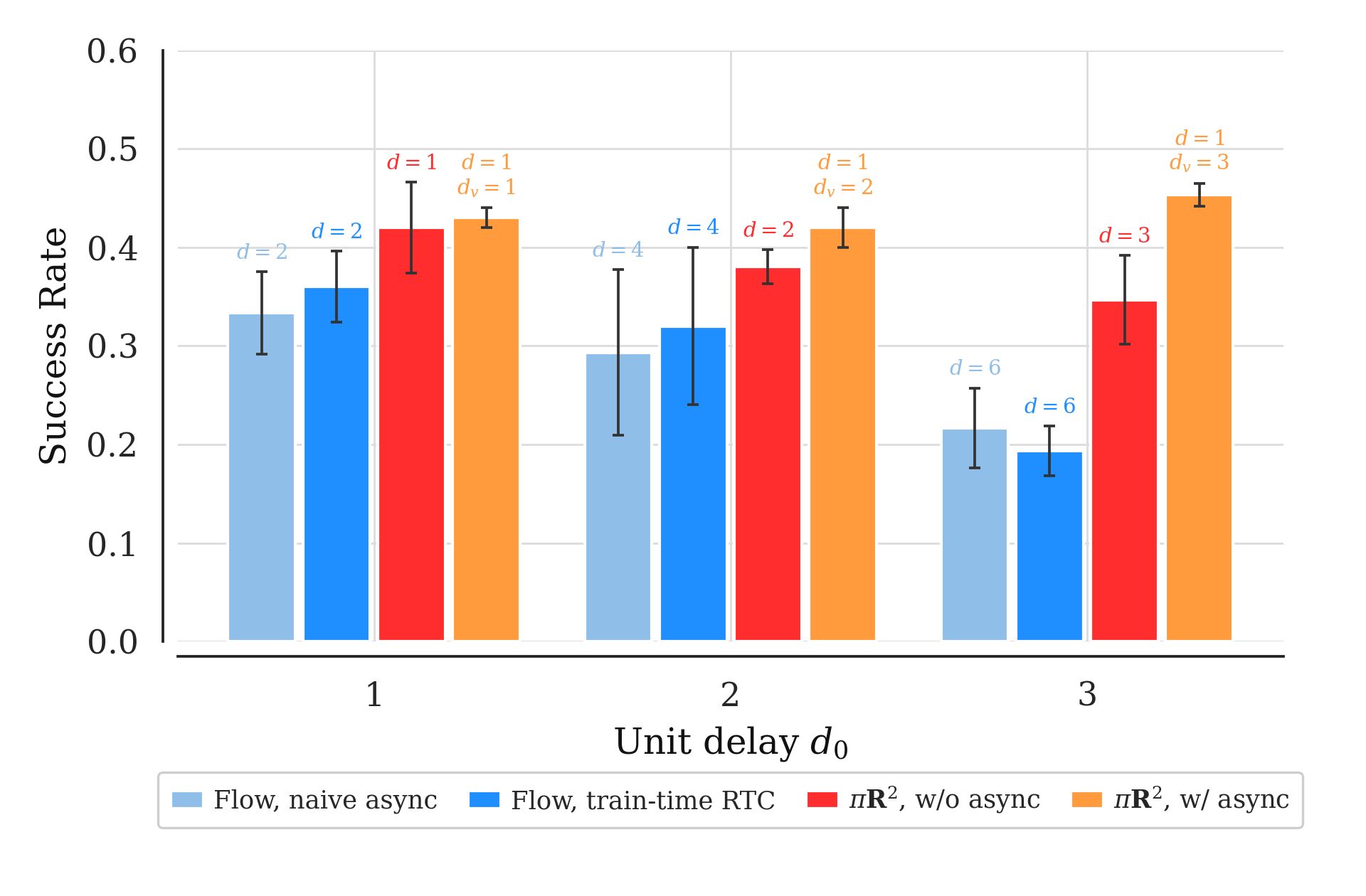}
    \end{subfigure}
    \vspace{-12pt}
    \caption{Simulation results. \textbf{Left.} Without any inference delay,  executing a smaller number of actions within the chunk benefits the performance for a flow matching policy.~\approach~also achieves the same performance via replanning every timestep, while the only last denoising step is conditioned on up-to-date state. \textbf{Right.} When there is an inference delay for each component,~\approach~reduces the effective delay $d$ by reducing the number of denoising steps and asynchronously processing the visual features. For each datapoint, $d$ indicates the effective delay of proprioception, while $d_v$ is visual delay. We train 3 policies with different seeds for each method and report the mean and std of the success rate.}
    \label{fig:simB}
    \vspace{-10pt}
  \end{figure}

  \subsubsection{Execution-horizon ($h$) sweep without inference delay}
  \label{sec:sim_zero}

  We first study the setting where inference delay $d=0$, to isolate two things: (i) the tradeoff between task performance and execution horizon $h$ (executing more actions open-loop
  misses the recent state), and (ii) whether \approach's amortized denoising, where each action's final denoising step conditions on the most recent observation, matches the reactivity of
  $h{=}1$ flow at a fraction of the per-call cost. At evaluation, standard flow matching uses $16$ denoising
  steps to predict the whole chunk, whereas~\approach~uses one denoising step and emits one action at a time. We sweep $h \in \{1, 2, 4, 8\}$ for standard flow; \approach\ emits one
  action per call ($h = 1$, $1$ NFE per call). 

\textbf{Results.}  Results are highlighted in Fig.~\ref{fig:simB} (left). Intuitively, the performance of standard flow degrades as $h$ grows, confirming the $h$ tradeoff for reactive tasks. Meanwhile,~\approach\ matches the flow configuration of standard flow $h \in \{1, 2\}$. This supports that leveraging fresh observation more frequently is beneficial for such reactive tasks, and amortizing the denoising budget
  across calls does not sacrifice quality, even though each call costs only one NFE. We next introduce realistic latency and test each deployment regime under that constraint.

 \subsubsection{Deployment under VLA-level inference delay}
  \label{sec:sim_delay}

To consider a realistic latency, we take into account GR00T-N1.7 computation cost (Sec.~\ref{sec:async-vlm}: vision/text VLM processing $\approx 60$~ms, $K{=}4$ NFE denoising $\approx 80$~ms) and define unit delay $d_0$ as the $K=4$ denoising cost, resulting in vision/text processing $\approx
  0.75\,d_0$ and one NFE $= d_0/4$. 

\textbf{Baselines.} We compare four methods: (i)~\textit{naive-async} runs the policy inference for predicting the new chunk while executing previous chunk. The inference does not take into account the actions being executed during inference. (ii)~\textit{train-time RTC}~\cite{black2025training} explicitly conditions the in-flight actions while predicting the new chunk, resulting in smooth chunk boundaries. (iii)~\approach\ without asynchronous fast/slow channel processing only applies latency-adaptive flow schedule (Sec.~\ref{sec:slope}) without asynchronously processing the fast and slow channel, and (iv)~\approach\ with asynchronous processing applies all proposed modifications.
  
  The effective per-call delays are then $1.75\,d_0$ (end-to-end inference:~\textit{naive-async} and~\textit{train-time RTC}), $1.0\,d_0$ (\approach\ w/o async), and $0.25\,d_0$ (\approach\ w async). Setting $d_0
  \in \{1, 2,3\}$ control ticks and ceil-rounding gives effective proprioception delay $d \in \{2,4,6\}/\{1,2,3\}/\{1,1,1\}$, respectively. For \approach\ with async processing, we additionally delay the object pose by $d_{vis} \in \{1,2,3\}.$ For all methods, we set execution horizon $h=d$ across all methods. 

\textbf{Results.} Fig.~\ref{fig:simB} (right) reports the mean over three training seeds (error bars show std); for each cell we pick the best epoch per seed. ~\approach\ w/ async wins at every $d_0$ ($0.43$, $0.42$, $0.45$), beating naive-async ($0.33$, $0.29$, $0.22$) and Train-time RTC ($0.36$, $0.32$, $0.19$) with margins that widen as delay grows. The reason is effective delay: baselines pay $d{=}\lceil 1.75\,d_0\rceil$ per call while~\approach\ pays $d{=}d_0$ (w/o async) or $d{=}1$ (w/ async). Async processing lets us hold the action delay at $1$ while only the visual delay $d_\mathrm{vis}$ grows, so performance degrades gracefully with $d_\mathrm{vis}$ -- supporting the view that vision provides coarse guidance while up-to-date proprioception drives manipulation.

\subsection{Real World Experiments}
\begin{figure}[t]
    \centering
    \includegraphics[width=0.95\linewidth]{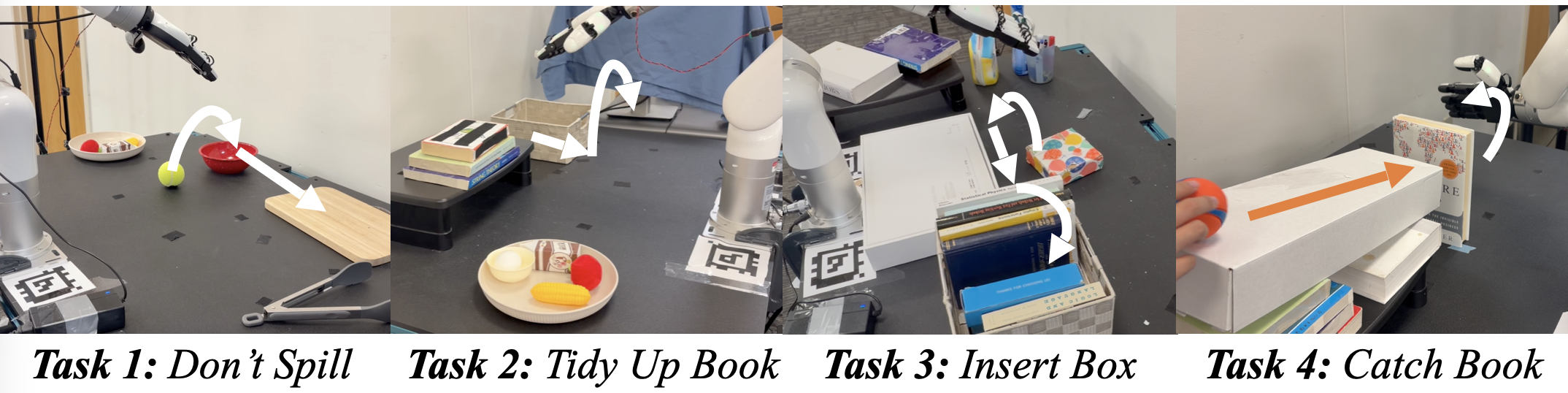}
    \vspace{-6pt}
    \caption{\textbf{Real-world manipulation tasks.} Arrows in each image show the desired task outcome.}
    \label{fig:real_tasks}
    \vspace{-10pt}
\end{figure}

\begin{table}[t]
  \centering
  \small
  \caption{\textbf{Real World Results.} (xArm6~+~XHand, fine-tuned GR00T N1.7). $d$ is the measured wall-clock inference delay in
  $25$-Hz control ticks ($1$ tick $\approx 40$~ms), which is the same as execution horizon $h$ for all methods other than synchronous inference. For each task we report success rate (SR) over $N$ trials and a partial-progress score (Prog) capturing the fraction of subgoals completed per episode.}

\label{tab:real}
  \begin{tabular}{lccccccc}
  \toprule
                                     &       \multicolumn{2}{c}{Don't Spill} & \multicolumn{2}{c}{Tidy up Book} & \multicolumn{2}{c}{Insert Box} & Catch Book \\
  \cmidrule(lr){2-3} \cmidrule(lr){4-5} \cmidrule(lr){6-7} \cmidrule(lr){8-8}
  Setting                              & SR & Prog & SR & Prog & SR & Prog & SR \\
  \midrule
  Flow, Synchronous ($h=10$)
      & 4/20 & 16/80 & 4/20 & 9/40  &11/20 & 56/80 & 4/20 \\
  Flow, Naive Async, dense, TE~\cite{zhao2023learning}
      & 7/20 & 30/80 & 7/20 & 15/40 & 12/20 & 61/80 & 2/20 \\
  Flow, Train-Time RTC~\cite{black2025training}
            & 9/20 & 45/80 & 8/20 & 18/40 & 10/20 & 53/80 & 5/20 \\
  \midrule
  \approach
      & \textbf{10/20} & \textbf{55/80}
      & \textbf{12/20} & \textbf{24/40}
      & \textbf{16/20} & \textbf{68/80}
      & \textbf{11/20} \\
  \bottomrule
  \end{tabular}
\end{table} 
  \label{sec:real}

  We now mirror the deployment study above on a real xArm6~+~XHand setup by fine-tuning  GR00T-N1.7 model~\cite{nvidia2025gr00tn1openfoundation} for dexterous manipulation. Demos are collected at $25$~Hz (capped by the teleoperation recording pipeline), so we report $d$ in $25$-Hz ticks ($1$ tick $\approx 40$~ms); We use RTX A5000 for both training and inference, resulting in $d=4\sim 5$ for baselines and $d=1\sim2$ for~\approach. We include arm and hand joint angles, along with hand per-joint torque and fingertip forces in the proprioceptive state.

\textbf{Tasks.}
We evaluate our method on four contact-rich dexterous, reactive manipulation tasks, shown in Fig.~\ref{fig:real_tasks}. \textit{Don't Spill} requires placing a ball into a bowl and moving the bowl onto a cutting board without dropping the ball.
\textit{Tidy Up Book} requires extracting a book from a pile, grasping it, and placing it into a basket.
\textit{Insert Box} requires pushing a box against a wall to stand it upright before inserting it between a row of books.
\textit{Catch Book} requires the robot reacting to the falling book (induced by the ball hitting the book) and grasping it without dropping it. 

All four tasks demand highly reactive behavior. Picking up the ball requires dynamic grasping, moving the bowl without tilting it relies on proprioceptive feedback, and both extracting a book from a pile and pushing a box upright require precise contact-rich interactions and reactive control. Lastly, catching a book requires the robot to quickly react so that the book stays within its palm. We collect 200, 300, 300, and 100 teleoperated demonstrations for each task, respectively.

\textbf{Baselines.}
  We compare \approach\ against three baselines, all fine-tuned from GR00T-N1.7 with same training budget. (1) \textit{Flow, Synchronous} stalls the robot for the full $d{=}4\sim 5$ ticks while
  inference runs, then executes the predicted chunk ($h = 10$). (2) \textit{Flow, Naive Async (dense, TE)}~\cite{zhao2024aloha} runs the policy asynchronously without conditioning on
  in-flight actions: \emph{dense} re-issues a query every $d$ ticks (whenever the previous call returns), and \emph{TE} (temporal ensembling)~\cite{zhao2023learning} for overlapping action predictions for
  the same timestep. (3) \textit{Flow, Train-Time RTC}~\cite{black2025training} additionally trains the policy to inpaint a clean $d$-action front, anchoring the new chunk to the executed
  history at the chunk boundary.

  \textbf{Results.}
  Table~\ref{tab:real} summarizes the experimental results. \approach\ with asynchronous vision--language inference operates near the per-control-tick limit ($d{=}1$ at $25$~Hz, with occasional increases to $d{=}2$ under network delays), whereas all flow-based baselines incur the full GR00T pipeline latency of $d \in {4,5}$. Synchronous inference produces jittery motion at chunk boundaries because the robot pauses while waiting for the next action chunk, leading to failures such as the ball rolling off the table. Naive asynchronous inference combined with temporal ensembling yields smoother motion but sacrifices execution precision. Train-Time RTC~\cite{black2025training} is the strongest baseline; however, \approach\ outperforms it across all metrics, with the largest gains on reactivity-critical tasks such as Tidy Up Book, Insert Box, and Catch Book, achieving approximately $20\sim30$\% absolute improvement. We observe that RTC also struggles to recover from failures because previously generated in-flight actions bias the policy toward continuing its recent motion, which reduces its ability to react promptly, particularly in Insert Box. Qualitatively, \approach\ produces smoother and more accurate motions, even during failure recovery. These results demonstrate the value of the real-time proprioceptive feedback enabled by \approach.

\begin{figure*}[t]
    \centering

    \includegraphics[width=0.96\textwidth]{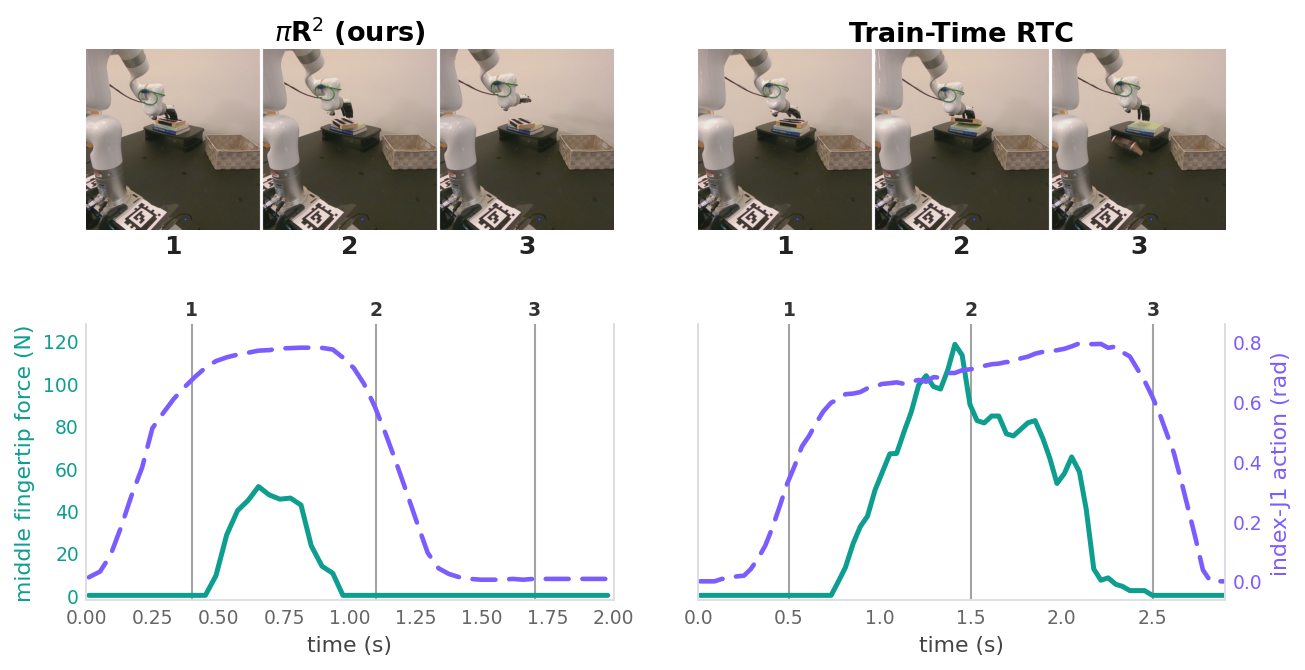}
    \vspace{-4pt}
    \caption{\textbf{\approach\ reacts to proprioception, while baselines run a stale plan} (\textit{Tidy Up Book}). Fingertip force (solid, left axis) and the emitted action (dashed, right axis) over time, for \approach\ and Train-Time RTC; numbered markers link plot times to the overhead frames above. \approach\ grips just enough ($\sim$50~N), while RTC reacts late and over-grips to $\sim$120~N, dropping the book.}
    \label{fig:qual}
\end{figure*}

\paragraph{Reactivity analysis.} Fig.~\ref{fig:qual} plots a fingertip contact force against the action each method emits over time on \textit{Tidy Up Book}. Because \approach\ refreshes proprioception at every denoising step, it modulates its grip from live force feedback and stops near $\sim$50~N, gripping just enough to reorient the book. Train-Time RTC instead commits to a stale plan and reacts late, so it keeps pushing down after contact, and its middle-finger force overshoots to $\sim$120~N, crushing the book. This force modulation is the mechanism behind our quantitative gains: reacting to contact as it happens rather than after the current chunk finishes. Tmaihe same pattern holds in all remaining tasks (Appendix~\ref{sec:appendix_reactivity}).

\section{Discussion}
  \label{sec:discussion}
  \vspace{-0.2cm}

  We presented \approach, a principled modification to existing VLA architectures that yields real-time, closed-loop policies through two orthogonal contributions: asynchronous
  vision/text processing that decouples the slow VLM backbone from the action loop, and a latency-adaptive flow schedule that absorbs variable inference delay via a per-position noise
  schedule parameterized by $d$. Together, they reduce per-call delay by up to $4\times$ while remaining compatible with any VLA action head. This enables \approach\ to handle highly
  reactive behaviors -- non-prehensile, contact-rich manipulation -- where baselines struggle, both in simulation and on real hardware.
  \vspace{-0.2cm}

\section*{Limitations}
  \label{sec:limit}
  \vspace{-5pt}
  Our approach has several limitations. First, we do not address sources of latency external to the model itself, such as communication delays between the inference server and the robot
  client. Second, we keep the base policy architecture unchanged; designing it to emphasize proprioceptive features more strongly (e.g., dedicated attention heads for proprioceptive
  tokens) could further amplify reactivity, and we leave this direction to future work.

\section*{Acknowledgements}
\label{sec:ack}
We appreciate the helpful discussions with CISCO, Tony Tao, Andrew Wang, Jason Liu, and Yuxuan Kuang. We also thank Yishu Li, Kallol Saha, and Soumojit Bhattacharya for helping real-world experiments. Lastly, we thank Hyeonwoo Kim for the feedback on website design. This work was supported by gift awards from CISCO, Google, and NSF Award IIS-2442282.
\bibliography{bib/reference}

\clearpage
\clearpage
\appendix

\section{Implementation Details}
\label{sec:sup_impl}

\subsection{Simulation}
\label{sec:sup_sim}

\paragraph{Task.}
The Leap Cube Reorientation task in MuJoCo Playground~\cite{zakka2025mujoco} requires a $16$-DoF Leap Hand to orient a cube without dropping it from the palm. We modify the original environment in two ways: we raise the control rate from $20$~Hz to $50$~Hz to enable more reactive control, and we restrict the goal distribution to four \emph{blue side up} yaw poses $\{0, {\pi/2}, \pi, {3\pi/2}\}$ rather than uniformly random goals. An episode succeeds if the cube reaches any of the four goals within $0.2$~rad in $600$ steps; we report the success rate over $100$ episodes.

\begin{figure}[ht]
  \centering
  \includegraphics[width=0.95\linewidth]{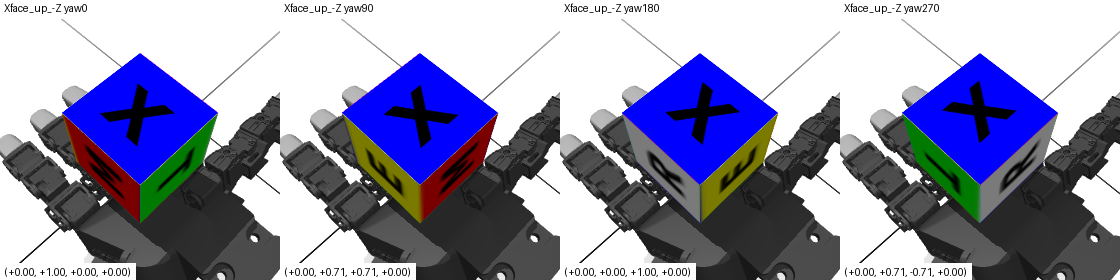}
  \caption{\textbf{Leap Cube Reorientation goals.} The Leap Hand grasps a cube and must rotate it until the blue face points up at one of four target yaw poses ($0^\circ, 90^\circ, 180^\circ, 270^\circ$). The bottom caption shows the corresponding goal quaternions $(w, x, y, z)$.}
  \label{fig:leap_overview}
\end{figure}

\paragraph{Dataset.}
Four PPO experts (different seeds) generate $50$ trajectories each, $200$ total at $50$~Hz. The simulator injects per-step sensing noise into the observation, sampled independently for each component: uniform noise in $[-0.05, 0.05]$~rad on joint angles, uniform noise in $[-0.02, 0.02]$~m on cube position, and Gaussian noise with scale $0.1$ on the cube orientation quaternion (the quaternion is re-normalized after perturbation). The recorded demonstrations preserve this noise, so the training distribution already includes realistic sensing noise on both the proprioceptive and vision-derived components.


\paragraph{Observation space.}
Each observation $\mathbf{o}_t$ concatenates:
\begin{itemize}[leftmargin=*,itemsep=1pt]
\item \textbf{Proprioception ($32$ dim)}: noisy joint angles ($16$) and per-joint tracking error -- noisy joint angle minus the most recent motor target ($16$).
\item \textbf{Vision-derived cube pose ($9$ dim)}: palm-to-cube position error ($3$) and absolute cube orientation in the world frame ($6$).
\end{itemize}
The $9$-dim vision-derived subset stands in for VLM features per Sec.~\ref{sec:async-vlm}; asynchronous-vision/text evaluations apply staleness only to this subset.

\paragraph{Action space.}
$16$-dim relative joint commands at $50$~Hz. Each action $\mathbf{a}_t \in [-1, 1]^{16}$ is scaled by $0.5$ and added to the current motor target.

\paragraph{Architecture.}
All variants share a Conditional U-Net 1D action head with chunk length $H{=}16$ and $n_\mathrm{obs}{=}2$. The U-Net uses FiLM~\cite{perez2018film} modulation by the noise level $\tau$ at every residual block; \approach\ replaces the shared FiLM projection with a per-position one -- one $(\gamma_p, \beta_p)$ pair per chunk position -- analogous to the per-position AdaLN change described in Sec.~\ref{sec:slope} for DiT-based heads. The rest of the architecture is identical across all variants.

\paragraph{Training schedule.}
At each batch we sample $d \sim \mathrm{Uniform}\{1, \dots, d_\mathrm{max}\}$ with $d_\mathrm{max}{=}5$, build $\boldsymbol{\tau}^{\star,d}$, fill the front-$d$ slots with ground-truth actions, and mask them out of the loss via $m_p = \mathbf{1}[p \ge d]$. With probability $0.2$ we instead train on a standard flow schedule (single $\tau \sim \mathrm{Uniform}[0,1]$ shared across positions, no mask), so the same network can also denoise a full chunk from pure noise. The standard-flow branch supplies the inference pass that initializes the buffer at episode start (Sec.~\ref{sec:slope}, warm-up). For the zero-delay sync curve in Sec.~\ref{sec:sim_zero}, we train a $d{=}0$-only variant ($d_\mathrm{max}{=}0$).

\paragraph{Slow-subset delay in simulation.}
The simulation policies are purely state-based: no image or language input. The $9$-dim vision-derived subset of the state vector (palm-to-cube position error and absolute cube orientation) plays the slow-channel role from Sec.~\ref{sec:async-vlm}. In this setting we find that neither delayed-visual-state sampling at training nor the learned delay embedding is necessary: the env already injects sensor noise on this subset at every step, so the policy is already tolerant to perturbed slow inputs out of the box. 

For all rows of Sec.~\ref{sec:sim_delay}, a single \approach\ checkpoint is used regardless of $d_\mathrm{vis}$: the synchronous slow-channel rows ($d_\mathrm{vis}{=}0$) and the asynchronous-vision/text rows ($d_\mathrm{vis}{>}0$) share the same network, and evaluation simply feeds a $d_\mathrm{vis}$-step-old version of the $9$-dim subset at test time.

\paragraph{Evaluation protocol.}
At inference, \approach\ performs one denoising step per policy call (each call outputs $n_{act}$ actions). Train-time RTC and standard flow runs $K{=}15$ denoising steps to predict the whole chunk. Each $(\mathrm{method}, d)$ cell in Sec.~\ref{sec:sim_delay} reports the mean success rate over $100$ episodes. Full sim hyperparameters are in Tab.~\ref{tab:hyperparameters_sim}.

\begin{table}[ht]
\centering
\small
\caption{\textbf{Simulation training and inference hyperparameters} for the Leap Cube Reorientation Conditional U-Net 1D policies. Values shown are \approach\ defaults; the right column lists deviations used by baselines. Empty = baseline same as \approach.}
\label{tab:hyperparameters_sim}
\begin{tabular}{lll}
\toprule
Parameter & \approach\ & Baseline deviations \\
\midrule
\multicolumn{3}{l}{\emph{Shared across all variants}} \\
\midrule
Chunk length $H$                        & \multicolumn{2}{l}{$16$} \\
Observation history $n_\mathrm{obs}$    & \multicolumn{2}{l}{$2$} \\
Control rate                            & \multicolumn{2}{l}{$50$~Hz} \\
Optimizer                               & \multicolumn{2}{l}{AdamW ($\beta{=}[0.95, 0.999]$)} \\
Learning rate (peak)                    & \multicolumn{2}{l}{$1{\times}10^{-4}$} \\
LR schedule                             & \multicolumn{2}{l}{cosine, $500$-step warm-up} \\
Weight decay                            & \multicolumn{2}{l}{$1{\times}10^{-6}$} \\
Gradient-norm clip                      & \multicolumn{2}{l}{$1.0$} \\
Batch size (global)                     & \multicolumn{2}{l}{$256$} \\
GPUs                                    & \multicolumn{2}{l}{$1\times$ A5000} \\
Training budget                         & \multicolumn{2}{l}{$800$ epochs} \\
\midrule
\multicolumn{3}{l}{\emph{Method-specific (noise schedule + inference)}} \\
\midrule
Train-time $d_\mathrm{max}$             & $5$ & $10$ (Train-time RTC); n/a (Flow) \\
Standard-flow warm-up prob.\ $\alpha$   & $0.2$ & n/a \\
Inference budget per call               & $1$ & $15$ (Train-time RTC, Flow)\\
\bottomrule
\end{tabular}
\end{table}

\paragraph{Baselines.}
All three variants share the dataset, optimizer, and training budget. \approach\ and train-time RTC additionally share the per-batch delay-sampling protocol: each batch draws $d \in \{0, \dots, d_\mathrm{max}\}$ from an exponentially-decaying distribution $p(d{=}k) \propto e^{-\alpha_d k}$ with $\alpha_d{=}1$ (smaller $d$ more likely following~\cite{black2025training}), and the front-$d$ positions are clamped to ground-truth actions and masked out of the loss. The three variants then differ only in:
\begin{itemize}[leftmargin=*,itemsep=1pt]
\item \textbf{Flow}: no $d$ sampling and no front clamp; a single shared noise level $\tau \sim \mathrm{Uniform}[0, 1]$ is applied to the entire chunk. This is standard chunked flow matching, evaluated at NFE${=}15$.
\item \textbf{Train-time RTC}~\cite{black2025training}: same per-batch $d$ sampling as \approach, but the noise schedule applies a single shared noise level $\tau \sim \mathrm{Uniform}[0, 1]$ to the back $H{-}d$ positions (no per-position structure). No standard-flow warm-up mixing, and larger $d_{max}=10$ as the effective delay is larger.
\item \textbf{\approach}: same per-batch $d$ sampling as train-time RTC, plus the $\alpha{=}0.2$ standard-flow warm-up branch (with probability $0.2$ the batch instead uses a single $\tau \sim \mathrm{Uniform}[0, 1]$ on all positions and no front clamp, so the same network can also initialize a buffer from pure noise at episode start). On the remaining $0.8$ of batches we apply the three-region staircase $\boldsymbol{\tau}^{\star,d}$ of Sec.~\ref{sec:slope}, and the U-Net uses per-position FiLM heads $\{(\gamma_p, \beta_p)\}_{p=0}^{H-1}$.
\end{itemize}

\subsection{Real-World}

\paragraph{Hardware and Workspace.}
\begin{wrapfigure}{r}{0.32\linewidth}
  \vspace{-1.2em}
  \centering
  \includegraphics[width=\linewidth]{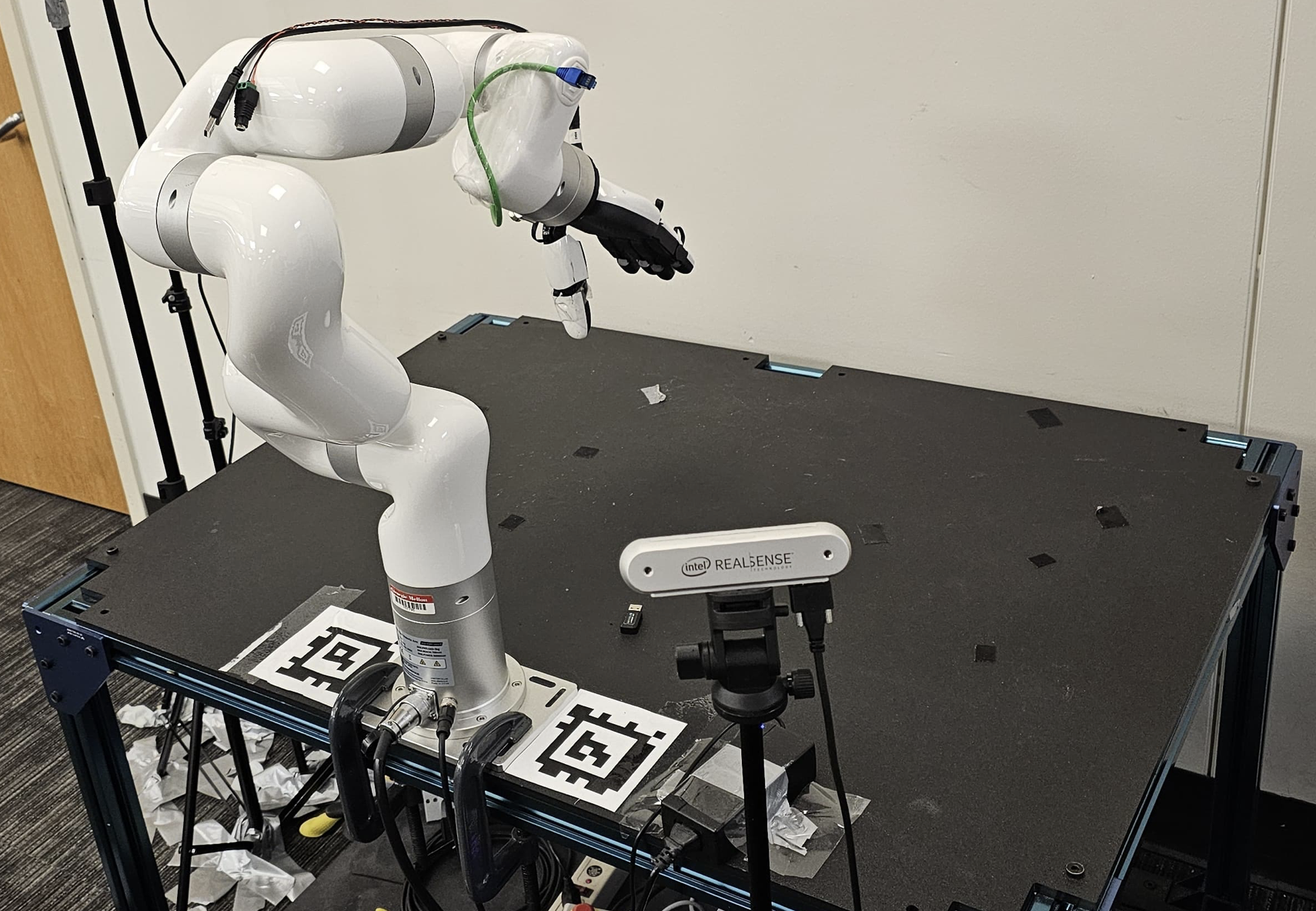}
  \caption{\textbf{Real-world workspace.} xArm6 + XHand with a single overhead RGB camera.}
  \label{fig:workspace}
  \vspace{-1.0em}
\end{wrapfigure}
A $6$-DoF xArm6 manipulator carries a $12$-DoF XHand at its wrist; a single overhead $640 \times 480$ RGB camera looks down on the workspace from above (Fig.~\ref{fig:workspace}). We fine-tune on $8 \times$ NVIDIA RTX A5000/A6000 ($24$/$48$~GB VRAM). Deployment uses a single RTX A5000 for the baselines.~\approach~uses $2 \times$ RTX A5000 so the slow VLM and the fast DiT action head do not share compute or memory bandwidth. Sharing one GPU between the two workers measurably inflates DiT latency and would conflate the measured $d$.

\paragraph{Control rate.}
Predicted actions run closed-loop at $25$~Hz at deployment ($T_\mathrm{ctrl} \approx 40$~ms per tick). Teleoperation captures demonstrations at the same rate, bottlenecked by the camera and the XHand--workstation USB link.

\paragraph{Base model.}
We fine-tune NVIDIA GR00T-N1.7~\cite{nvidia2025gr00tn1openfoundation}. The Qwen-VL backbone and Eagle-2 vision encoder remain \emph{frozen}; only the action head -- state/action projectors, the position embedding, and the DiT -- is trained. The only architectural change to the action head replaces the DiT's shared AdaLN modulation with the per-position $(\gamma_p, \beta_p)$ heads of Sec.~\ref{sec:slope}.

\paragraph{Observation space.}
Each call to the policy receives:
\begin{itemize}[leftmargin=*,itemsep=1pt]

\item \textbf{Proprioception} ($45$ dim): $6$ xArm6 joint angles, $12$ XHand joint angles (the actively-controlled finger DoFs), $12$ corresponding per-joint torques, and $15$ fingertip-force values ($3$-axis force from each of $5$ sensorized fingertips). 

\item \textbf{Vision and language}: one $640 \times 480$ RGB image from the overhead camera, encoded by Eagle-2 into vision tokens, concatenated with the task prompt (Tab.~\ref{tab:real_tasks_prompt}) at the VLM backbone input.
\end{itemize}

\paragraph{Action space and low-level drivers.}
Each policy call emits a chunk of $H{=}50$ absolute joint-position targets ($6$ xArm6 + $12$ XHand) at $25$~Hz, spanning $2$~s of motion. We send one target per control tick to each robot's driver in position-control mode. The arms' onboard firmware interpolates between consecutive commands at their internal servo rates(Tab.~\ref{tab:driver_settings}).

\begin{table}[ht]
\centering
\small
\caption{\textbf{Low-level driver settings.} Both robots run in position-only mode at $25$~Hz outbound from the policy; the onboard firmware interpolates between targets at its internal rate.}
\label{tab:driver_settings}
\begin{tabular}{lll}
\toprule
& xArm6 & XHand \\
\midrule
Driver mode             & \texttt{mode=1} (streaming servo) & RS485, position mode \\
Command call            & \texttt{set\_servo\_angle\_j}     & finger \texttt{set\_target} \\
Bus / link              & TCP                                & RS485 \\
Onboard control         & API (PID + interp.)    & per-joint PID, $k_p{=}100,\ k_i{=}0,\ k_d{=}0$ \\
Safety limits           & collision sensitivity $= 2$        & torque limit $= 300$ \\
\bottomrule
\end{tabular}
\end{table}

\paragraph{Data collection.}
We teleoperate the robot setup to collect $200$ demonstrations for \textit{Don't Spill}, $300$ for \textit{Tidy Up Book} and \textit{Insert Box}, and $100$ for \textit{Catch Book}, randomizing the target object's initial pose.

\begin{table}[ht]
\centering
\small
\caption{\textbf{Real-world task prompts.} Exact language input passed to GR00T-N1.7 at both training and deployment.}
\label{tab:real_tasks_prompt}
\begin{tabular}{ll}
\toprule
Task & Language prompt \\
\midrule
Don't Spill   & \texttt{put the ball in the bowl and put them on the cutting board} \\
Tidy Up Book  & \texttt{put the books in the basket} \\
Insert Box  & \texttt{put the box in the basket} \\
Catch Book  & \texttt{catch the book} \\
\bottomrule
\end{tabular}
\end{table}

\paragraph{Training the asynchronous vision/text mechanism.}
The real-world \approach\ training explicitly exercises the asynchronous-vision/text mechanism of Sec.~\ref{sec:async-vlm}. Each batch samples a vision delay $d_\mathrm{vis} \sim \mathrm{Uniform}\{0, \dots, d_\mathrm{vis}^{\max}\}$ with $d_\mathrm{vis}^{\max}{=}5$ ticks ($200$~ms at $25$~Hz); the policy then receives the image recorded $d_\mathrm{vis}$ ticks earlier (i.e., the corresponding past frame in the demonstration), while proprioception stays current. For each batch, the integer $d_\mathrm{vis} \in \{0, \dots, d_\mathrm{vis}^{\max}\}$ indexes a learned embedding $e(d_\mathrm{vis})$ from a $(d_\mathrm{vis}^{\max}{+}1)$-entry lookup table. The DiT adds $e(d_\mathrm{vis})$ to its action-token features (broadcast across the $H$ chunk positions), so the action head conditions on both the cached visual feature and its age. We zero-initialize the lookup table so that the untrained checkpoint reproduces the no-delay variant exactly. At deployment, the measured wall-clock vision latency passes through the same embedding before each DiT call.

\paragraph{Training hyperparameters.}
We fine-tune the GR00T-N1.7 action head (projectors + DiT) and keep the VLM backbone and vision encoder frozen. Full optimizer settings, batch size, and step budgets are in Tab.~\ref{tab:hyperparameters_real}. The per-task budgets ($40{,}000$ steps for \textit{Don't Spill}, $28{,}000$ for \textit{Tidy Up Book}, $49{,}000$ for \textit{Insert Box}, $4,550$ for \textit{Catch Book}) correspond to approximately $200/100/100/100$ epochs over each dataset. The per-position AdaLN parameters $\{(\gamma_p, \beta_p)\}_{p=0}^{H-1}$ initialize from the pretrained shared pair, so each head starts from a position-uniform schedule and gradually specializes during fine-tuning. All four rows in Tab.~\ref{tab:real} fine-tune from the same GR00T-N1.7 checkpoint with identical data and budget; they differ only in the noise schedule and (for \approach) the AdaLN modification. The train-time RTC baseline samples its per-batch delay from $d \in \{0, \dots, 10\}$ (i.e., $d_\mathrm{max}{=}10$) rather than the $d_\mathrm{max}{=}5$ used in \approach, as~\approach is separating the delay into two parts while RTC treats it as a whole. Full hyperparameters are in Tab.~\ref{tab:hyperparameters_real}.

\paragraph{Evaluation protocol.}
Each $(\mathrm{method}, \mathrm{task})$ cell runs $N{=}20$ trials with randomized initial object placement. Success rate (SR) counts whole-task completion within the episode time limit (30 seconds). The progress score (Prog) decomposes each task into sub-goals and reports the per-trial fraction completed:
\begin{itemize}[leftmargin=*,itemsep=1pt]
\item \textit{Don't Spill} -- $4$ sub-goals: pick up ball, place ball in bowl, lift bowl off the table, place bowl on the cutting board.
\item \textit{Tidy Up Book} -- $2$ sub-goals: pull a book free from the pile, place it inside the basket.
\item \textit{Insert Box} -- $4$ sub-goals: push the box towards the wall, make it stand, grasp, and insert between books.
\item \textit{Catch Book} -- $1$ sub-goal: grasp the falling book and pick it up.

\end{itemize}

\paragraph{Deployment query modes.}
The four rows of Tab.~\ref{tab:real} differ in how the action-head worker queries the policy:
\begin{itemize}[leftmargin=*,itemsep=0pt]
\item \textit{Flow, Synchronous}: the main loop blocks until each call returns; the robot freezes during inference.
\item \textit{Flow, Naive Async, dense, ensemble}: the action worker queries back-to-back and combines overlapping chunks with the temporal-ensembling weights of~\citet{zhao2023learning}.
\item \textit{Train-time RTC} and \textbf{\approach}: the action worker queries back-to-back; the new chunk replaces the active one at the next swap boundary. The actions executed during policy call is given as input.~\approach~additionally spawns a VLM-cache worker on a separate GPU that refreshes the cached visual features in the background.

\end{itemize}

\begin{table}[ht]
\centering
\small
\caption{\textbf{Real-world training and inference hyperparameters} for the GR00T-N1.7 fine-tunes. Values shown are \approach\ defaults; the right column lists deviations used by baselines. Empty = baseline same as \approach.}
\label{tab:hyperparameters_real}
\begin{tabular}{lll}
\toprule
Parameter & \approach\ & Baseline deviations \\
\midrule
\multicolumn{3}{l}{\emph{Shared across all variants}} \\
\midrule
Chunk length $H$                        & \multicolumn{2}{l}{$50$} \\
Observation history $n_\mathrm{obs}$    & \multicolumn{2}{l}{$1$} \\
Control rate                            & \multicolumn{2}{l}{$25$~Hz} \\
Optimizer                               & \multicolumn{2}{l}{fused AdamW} \\
Learning rate (peak)                    & \multicolumn{2}{l}{$1{\times}10^{-4}$} \\
LR schedule                             & \multicolumn{2}{l}{cosine, $5\%$ warm-up} \\
Weight decay                            & \multicolumn{2}{l}{$1{\times}10^{-5}$} \\
Gradient-norm clip                      & \multicolumn{2}{l}{$1.0$} \\
Batch size                     & \multicolumn{2}{l}{$512$} \\
Precision                               & \multicolumn{2}{l}{bf16 + tf32} \\
GPUs                                    & \multicolumn{2}{l}{$8\times$ A5000/A6000} \\
Training budget                         & \multicolumn{2}{l}{$100$ epoch (\textit{Tidy Up Book} / \textit{Insert Box} \ \textit{Catch Book})} / $200$ epoch (\textit{Don't Spill})
\\
\midrule
\multicolumn{3}{l}{\emph{Method-specific (noise schedule + delay + inference)}} \\
\midrule
Train-time $d_\mathrm{max}$             & $5$ & $10$ (Train-time RTC); n/a (Flow) \\
Train-time $d_\mathrm{vis}^{\max}$      & $5$ & n/a \\
Delay-embedding lookup size             & $6$ entries $\to$ DiT hidden dim & n/a \\
Standard-flow warm-up prob.\ $\alpha$   & $0.2$ & $0$ (Flow, Train-time RTC) \\
NFE per call                            & $1$ (one sub-chunk) & $4$ (Flow, Train-time RTC; full chunk) \\
\end{tabular}
\end{table}

\subsection{Training and Inference Algorithms}
\label{sec:sup_algo}

We give pseudocode for the per-batch training step (Alg.~\ref{alg:train}) and the deployment-time inference loop (Alg.~\ref{alg:infer}). The inference loop is the loop implemented in our deployment script; the wall-clock measurement that auto-derives $d$ uses a rolling window of the most recent query times.

\begin{algorithm}[ht]
\caption{\approach\ training step. Each batch samples a delay $d$, builds the staircase $\boldsymbol{\tau}^{\star,d}$, masks the front, applies jitter, and (for real-world training) additionally delays the slow channel.}
\label{alg:train}
\begin{algorithmic}[1]
\Require Action chunk $\mathbf{a}_{0:H-1}$, observation $(\mathbf{o}^\mathrm{fast}_t, \mathbf{o}^\mathrm{slow}_{t:t-T_\mathrm{slow}+1})$, model $v_\theta$
\Require Hyperparams $d_\mathrm{max}, d_\mathrm{vis}^{\max}, j, \alpha$
\State $u \sim \mathrm{Uniform}[0, 1]$
\If{$u < \alpha$}
  \Comment{warm-up branch: standard flow}
  \State $t \sim \mathrm{Uniform}[0, 1]$;\ $\tau_p \leftarrow t,\ \forall p$
  \State $\mathbf{x}_{\tau,p} \leftarrow (1-\tau_p) \boldsymbol{\epsilon}_p + \tau_p \mathbf{a}_p$
  \State $m_p \leftarrow 1,\ \forall p$
  \Comment{all positions contribute to loss}
\Else
  \Comment{staircase branch}
  \State $d \sim \mathrm{Uniform}\{1, \dots, d_\mathrm{max}\}$
  \State Build $\boldsymbol{\tau}^{\star,d}$ as in Sec.~\ref{sec:slope}\Comment{three-region staircase}
  \State $\delta_p \sim \mathrm{Uniform}[-j, j]$;\ $\tau_p \leftarrow \mathrm{clip}(\tau^{\star,d}_p + \delta_p, 0, 1)$
  \State $\mathbf{x}_{\tau,p} \leftarrow (1-\tau_p) \boldsymbol{\epsilon}_p + \tau_p \mathbf{a}_p$
  \State $\mathbf{x}_{\tau,p} \leftarrow \mathbf{a}_p$ if $p < d$\Comment{front $d$ slots are ground-truth (inpaint conditioning)}
  \State $m_p \leftarrow \mathbf{1}[p \ge d]$
\EndIf
\State $d_\mathrm{vis} \sim \mathrm{Uniform}\{0, \dots, d_\mathrm{vis}^{\max}\}$\Comment{slow-channel staleness (real-world only)}
\State $\mathbf{o}^\mathrm{slow} \leftarrow \mathbf{o}^\mathrm{slow}_{t-d_\mathrm{vis}}$;\ append $e(d_\mathrm{vis})$ to slow representation
\State $\hat{\mathbf{v}}_p \leftarrow v_\theta(\mathbf{x}_\tau, \boldsymbol{\tau}, \mathbf{o}^\mathrm{fast}_t, \mathbf{o}^\mathrm{slow})_p$
\State $\mathcal{L} \leftarrow \sum_{p=0}^{H-1} m_p \, \| \hat{\mathbf{v}}_p - (\mathbf{a}_p - \boldsymbol{\epsilon}_p) \|^2$
\State Update $\theta$ on $\mathcal{L}$
\end{algorithmic}
\end{algorithm}

\begin{algorithm}[ht]
\caption{\approach\ inference loop on the robot. The action head runs continuously with fresh proprioception and a cached slow feature; the VLM runs asynchronously in a background thread; the per-call delay $d$ is auto-derived from a rolling measurement of action-head wall-clock latency.}
\label{alg:infer}
\begin{algorithmic}[1]
\Require Policy $v_\theta$, observation source, robot controller, control period $T_\mathrm{ctrl}$
\Require Rolling window $W$ of recent query times (e.g.\ $W{=}20$)
\State $\mathbf{C} \leftarrow \texttt{WarmStart}()$\Comment{Sec.~\ref{sec:slope}: full standard-flow inference, denoise all $H$ positions}
\State $\mathbf{C} \leftarrow$ re-noise $\mathbf{C}$ to match $\boldsymbol{\tau}^{\star,d_0}$ for initial estimate $d_0$
\State $i \leftarrow 0$\Comment{chunk index}
\State queue $Q \leftarrow \emptyset$\Comment{rolling latency window}
\State \textbf{Spawn} \texttt{VLM\_Worker} thread (continuously updates cached slow feature)
\State \textbf{Spawn} \texttt{Action\_Worker} thread (continuously queries action head)
\Loop\Comment{one iteration per control tick ($25$~Hz)}
  \State $t_0 \leftarrow $ current wall-clock time
  \State Read fresh proprioception $\mathbf{o}^\mathrm{fast}_t$ from robot sensors
  \State Publish $\mathbf{o}^\mathrm{fast}_t$ to shared state for workers
  \If{new chunk $\mathbf{C}_\mathrm{new}$ is available from \texttt{Action\_Worker}}
    \State $\mathbf{C} \leftarrow \mathbf{C}_\mathrm{new}$;\ \ $i \leftarrow d$\Comment{swap and skip past in-flight front actions}
  \EndIf
  \State Send action $\mathbf{C}_i$ to robot;\ \ $i \leftarrow i + 1$
  \State Sleep until $t_0 + T_\mathrm{ctrl}$
\EndLoop
\Statex
\Procedure{Action\_Worker}{} \Comment{background thread; one denoising step per call}
\Loop
  \State Snapshot $(\mathbf{o}^\mathrm{fast}_t, t_\mathrm{state})$ and the cached $(\mathbf{o}^\mathrm{slow}_\mathrm{cached}, t_\mathrm{image})$
  \State $d_\mathrm{vis} \leftarrow \mathrm{round}\big((t_\mathrm{state} - t_\mathrm{image}) / T_\mathrm{ctrl}\big)$
  \Comment{age of cached slow feature in control ticks}
  \State $d \leftarrow \max\big(1,\ \mathrm{round}(\mathrm{mean}(Q) / T_\mathrm{ctrl})\big)$\Comment{auto-derived from rolling window}
  \State Snapshot front-$d$ inpaint: $\mathbf{a}^\mathrm{inflight}_{0:d} \leftarrow \mathbf{C}_{i:i+d}$
  \State $q_0 \leftarrow $ wall-clock
  \State $\mathbf{C}_\mathrm{new} \leftarrow v_\theta.\texttt{euler\_step}(\mathbf{a}^\mathrm{inflight}, \mathbf{o}^\mathrm{fast}, \mathbf{o}^\mathrm{slow}_\mathrm{cached}, e(d_\mathrm{vis}))$
  \Comment{one NFE; per-position $\Delta\tau_p$ as in Eq.~(\ref{eq:euler-step})}
  \State Push $\mathrm{wall\text{-}clock} - q_0$ into $Q$
  \State Hand off $\mathbf{C}_\mathrm{new}$ to main loop
\EndLoop
\EndProcedure
\Statex
\Procedure{VLM\_Worker}{} \Comment{background thread; runs vision/text forward asynchronously}
\Loop
  \State Read fresh image and language;\ $t_\mathrm{image} \leftarrow$ image capture time
  \State $\mathbf{o}^\mathrm{slow}_\mathrm{cached} \leftarrow $ VLM forward (image, language)
  \State Atomically update cache with $(\mathbf{o}^\mathrm{slow}_\mathrm{cached}, t_\mathrm{image})$
\EndLoop
\EndProcedure
\end{algorithmic}
\end{algorithm}

\subsection{Additional Reactivity Analysis}
\label{sec:appendix_reactivity}
\begin{figure*}[t]
    \centering
    \includegraphics[width=0.95\linewidth]{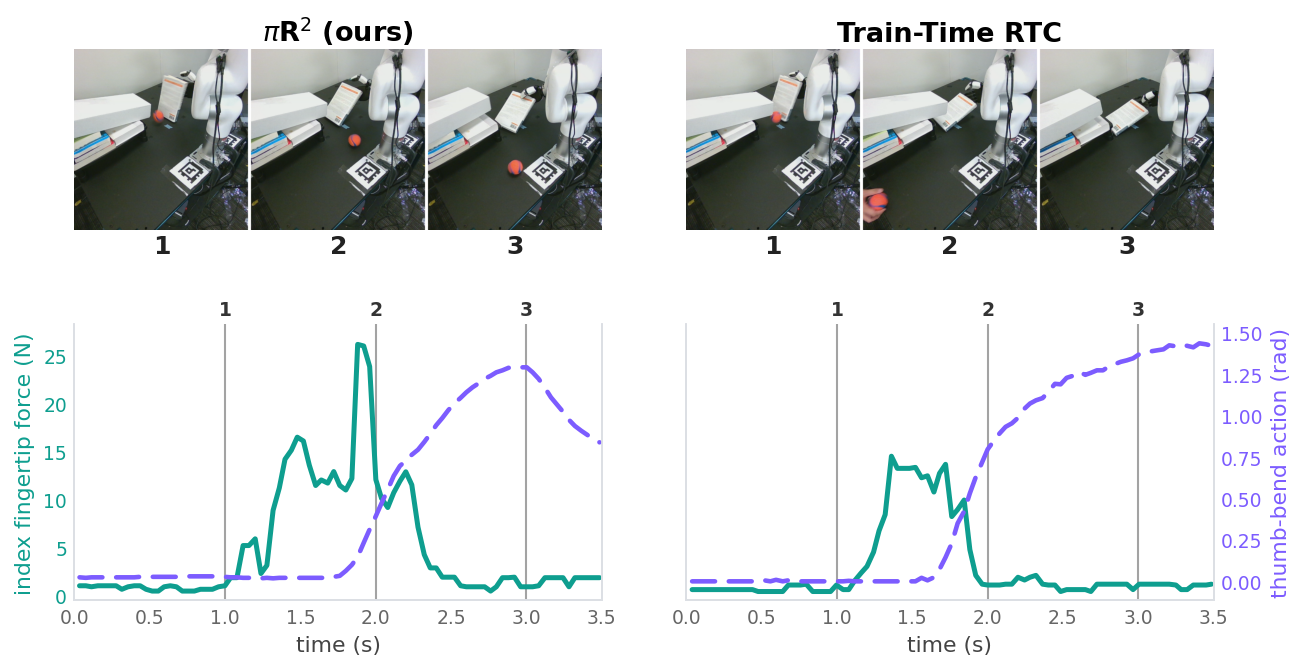}\\[4pt]
    \includegraphics[width=0.95\linewidth]{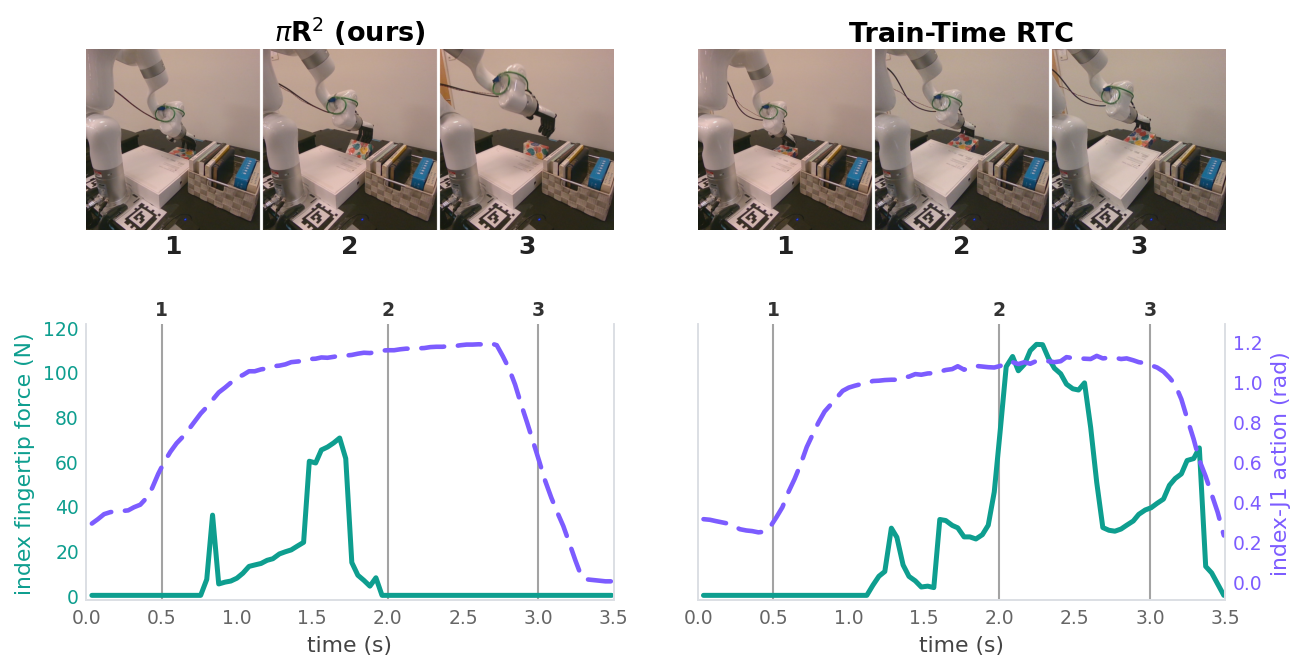}\\[4pt]
    \includegraphics[width=0.95\linewidth]{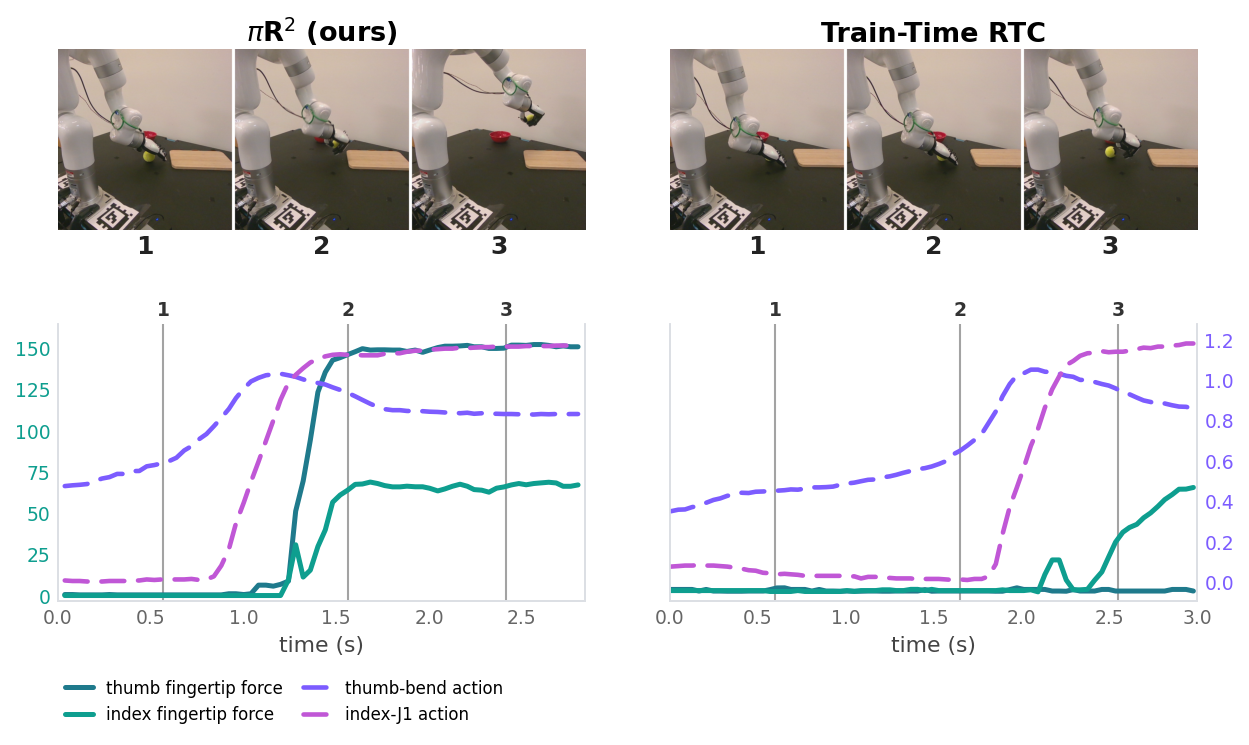}
    \caption{\textbf{Reactivity on the remaining tasks.} Fingertip force (solid, left axis) and the emitted action (dashed, right axis) over time for \approach\ and Train-Time RTC, with overhead frames at the marked times. Top: \textit{Catch Book}. Middle: \textit{Insert Box}. Bottom: \textit{Don't Spill}.}
    \label{fig:appendix_reactivity}
\end{figure*}

The main-text reactivity comparison (Fig.~\ref{fig:qual}) is on \textit{Tidy Up Book}; the same pattern holds across the other tasks (Fig.~\ref{fig:appendix_reactivity}). On \textit{Catch Book}, \approach\ closes on the force spike as the book lands, whereas Train-Time RTC reacts too late and the book slips. On \textit{Insert Box}, RTC reacts late to the force feedback and its index force overshoots out of distribution, while \approach\ stays controlled. On \textit{Don't Spill}, RTC's thumb never contacts the ball (thumb force $\approx 0$) yet it keeps going, while \approach\ grasps it with both fingers.

\end{document}